\pdfoutput=1
\documentclass[12pt, a4paper, oneside]{article} 

\usepackage{amsmath, amssymb, graphicx}
\usepackage[colorlinks=true, citecolor=blue, linkcolor=black]{hyperref} 

\usepackage{caption}
\usepackage{float}
\usepackage{cite}
\usepackage{array}
\usepackage{booktabs}
\usepackage{titlesec}
\usepackage{newtxtext, newtxmath} 

\usepackage{tocloft}  
\titleformat{\paragraph}
{\normalfont\normalsize\bfseries}{\theparagraph}{1em}{}
\titlespacing*{\paragraph}
{0pt}{3.25ex plus 1ex minus .2ex}{1.5ex plus .2ex}

\footnotetext{This research was supported by the China Scholarship Council (CSC).}

\title{
	\bfseries OCT Data is All You Need: How Vision Transformers with and without Pre-training Benefit Imaging \\ 
	\normalfont\normalsize \vspace{0.5em} 
}

\author{
	Zihao Han\textsuperscript{*}, 
	Philippe De Wilde \\
	University of Kent, \\
	Canterbury CT2 7NZ, United Kingdom \\
	\textsuperscript{*}Corresponding author: 
	\protect\href{mailto:zh219@kent.ac.uk}{zh219@kent.ac.uk}
}

\date{} 

\usepackage{titlesec}
\titleformat{\section}{\normalfont\Large\bfseries}{\thesection}{1em}{}
\titleformat{\subsection}{\normalfont\large\bfseries}{\thesubsection}{1em}{}
\titleformat{\subsubsection}{\normalfont\normalsize\bfseries}{\thesubsubsection}{1em}{}

\begin{document}
	\maketitle

	\tableofcontents
	\thispagestyle{empty}
	\newpage
	
	\section*{Abstract}
	\addcontentsline{toc}{section}{Abstract}
	\setcounter{page}{1}
	\noindent
	Optical Coherence Tomography (OCT) provides high-resolution cross-sectional images useful for diagnosing various diseases, but their distinct characteristics from natural images raise questions about whether large-scale pre-training on datasets like ImageNet is always beneficial. In this paper, we investigate the impact of ImageNet-based pre-training on Vision Transformer (ViT) performance for OCT image classification across different dataset sizes. Our experiments cover four-category retinal pathologies (CNV, DME, Drusen, Normal). Results suggest that while pre-training can accelerate convergence and potentially offer better performance in smaller datasets, training from scratch may achieve comparable or even superior accuracy when sufficient OCT data is available. Our findings highlight the importance of matching domain characteristics in pre-training and call for further study on large-scale OCT-specific pre-training.
	
	\vspace{1em}
	\noindent
	\textbf{Keywords:} Optical Coherence Tomography (OCT), Vision Transformer (ViT), OCT Image Classification, Transfer Learning
	
	\newpage  
	\clearpage  
	\section{Introduction}
	\label{sec:intro}

	Transformers, originally proposed by Vaswani et al. \cite{8Vaswani2017}, have transformed natural language processing (NLP) by introducing a self-attention mechanism capable of capturing long-range dependencies. This architecture has been successfully extended to computer vision (CV) through the Vision Transformer (ViT) framework \cite{10Dosovitskiy2020}, where images are divided into patches and processed as sequences. ViT has demonstrated competitive performance against convolutional neural networks (CNNs) across various image classification tasks, showcasing its potential to handle complex visual data.
	
	Optical Coherence Tomography (OCT) is a crucial imaging modality that provides high-resolution, cross-sectional views of the retina, enabling early detection and monitoring of diseases such as diabetic macular edema (DME) and choroidal neovascularization (CNV) \cite{35Nassif2004,36Pircher2017}. Compared to natural images, OCT scans exhibit distinct contrast, texture, and noise distributions \cite{37Ran2021,16Kermany2018}, making direct application of algorithms developed for natural-image domains less straightforward. Furthermore, annotated medical datasets are often limited in scale and expensive to acquire, increasing the complexity of training high-capacity models for robust diagnostic performance.
	
	As a common strategy to tackle data scarcity, researchers typically leverage pre-training on large-scale natural image datasets like ImageNet to initialize deep models before fine-tuning them on medical tasks \cite{27Raghu2019,28He2019}. While such pre-trained weights can improve convergence speed and sometimes enhance generalization, recent work highlights that a large domain gap(i.e., fundamental differences in imaging physics, structural representations, and data distributions between OCT and natural images)may reduce or even negate these benefits in specialized tasks such as OCT classification \cite{27Raghu2019,28He2019}. Indeed, the lack of OCT-specific pre-trained models often forces practitioners to rely on ImageNet-based weights, potentially underexploiting the unique structural features of OCT scans.
	
	In this paper, we systematically investigate whether pre-training on ImageNet21K is truly advantageous—or even necessary—for OCT classification. We focus on a four-category retinal pathology dataset (CNV, DME, Drusen, Normal) \cite{16Kermany2018} and compare performance between \textbf{ViT (Pre-trained)} and \textbf{ViT (Scratch)} under different data scale conditions. Our results show that while ImageNet pre-training can accelerate early convergence in small datasets, training from scratch can achieve comparable or superior performance when sufficient in-domain OCT data is available. These findings underscore the importance of domain alignment in transfer learning and motivate future exploration into large-scale OCT-specific pre-training or self-supervised methods tailored to medical imaging.

	\section{Related Work}
	\label{sec:related}
	
	Recent years have witnessed substantial progress in medical imaging analysis, driven by the convergence of deep learning algorithms, large-scale computing resources, and continuously expanding datasets. Table~\ref{tab:relatedsummary} provides a concise overview of representative methods in this domain, illustrating the historical dominance of CNNs, emerging Transformer-based approaches, and ongoing debates surrounding transfer learning strategies.
	
	\begin{table}[ht]
		\centering
		\caption{Representative Related Works in Medical Imaging and Transfer Learning}
		\label{tab:relatedsummary}
		\begin{tabular}{p{3.2cm}p{3.6cm}p{5cm}p{1.2cm}}
			\toprule
			\textbf{Approach / Model} 
			& \textbf{Dataset / Modality} 
			& \textbf{Key Findings} 
			& \textbf{Refs} \\
			\midrule
			\textbf{CNN for OCT lesions}  
			& Public OCT datasets  
			& High accuracy but data-hungry; limited long-range context modeling  
			& \cite{30Esteva2017,16Kermany2018} \\
			
			\textbf{Data augmentation \& GAN}  
			& Liver lesion, etc.  
			& Synthetic data helps mitigate class imbalance and small-sample issues  
			& \cite{31Frid-Adar2018} \\
			
			\textbf{Transformer (NLP to CV)}
			& ImageNet, COCO  
			& Global attention outperforms or competes with CNN, yet needs large data  
			& \cite{12Wang2018,10Dosovitskiy2020} \\
			
			\textbf{ViT in medical imaging}  
			& Brain MRI, CT  
			& Hybrid or specialized Transformer variants proposed for better data efficiency  
			& \cite{32Touvron2021,33Zhou2022,40Azizi2021} \\
			
			\textbf{OCT Domain Gap}  
			& Retinal OCT, OCTA  
			& Domain-specific structures hamper direct use of standard CNN/ViT models  
			& \cite{35Nassif2004,38Abramoff2010,37Ran2021,41Ran2021} \\
			
			\textbf{ImageNet pre-training}  
			& Various medical tasks  
			& Speeds up convergence; domain gap can lessen benefits  
			& \cite{34Deng2009,28He2019} \\
			
			\textbf{Scratch vs. Pre-trained}  
			& Medical tasks (general)  
			& Under certain conditions, scratch training can match or exceed pre-trained  
			& \cite{28He2019,27Raghu2019} \\
			\bottomrule
		\end{tabular}
	\end{table}
	
	\subsection{CNNs in Medical Imaging}
	Deep convolutional neural networks (CNNs) have been a approach for various medical imaging tasks, including lesion detection in CT/MRI \cite{29Litjens2017} and automated diagnosis in OCT \cite{30Esteva2017,16Kermany2018}. By leveraging hierarchical feature extraction, CNNs can yield high accuracy when provided with sufficiently large datasets. However, many medical image collections remain limited or imbalanced, increasing the risk of overfitting \cite{37Ran2021}. To counter this, researchers often employ data augmentation or generative adversarial networks (GAN) for synthetic data generation \cite{31Frid-Adar2018}, as well as transfer learning from ImageNet-pretrained models.
	
	In particular, Optical Coherence Tomography (OCT) is a crucial modality for ophthalmic diagnosis, enabling cross-sectional visualization of the retina that distinctly differs from natural images in contrast, texture, and noise characteristics \cite{35Nassif2004,38Abramoff2010}. Although CNN-based pipelines have shown success on OCT datasets \cite{16Kermany2018,41Ran2021}, the unique domain-specific features of retinal scans can limit the direct transferability of features learned from natural-image datasets.
	
	\subsection{Vision Transformers}
	Transformers, originally successful in NLP \cite{12Wang2018}, were introduced to computer vision through the Vision Transformer (ViT) \cite{10Dosovitskiy2020}, achieving performance on par with or surpassing CNNs in large-scale settings such as ImageNet. Unlike convolution-based models, ViT treats an image as a sequence of patches, allowing self-attention to capture long-range dependencies. This global modeling property can be particularly beneficial for identifying subtle lesions or anatomical variations in medical scans. 
	
	Nevertheless, ViT’s data-hungry nature poses challenges for medical imaging, where annotated datasets are often limited. To address these issues, hybrid or specialized Transformer architectures have been proposed, combining convolutional layers with attention blocks to strike a balance between local and global feature learning \cite{33Zhou2022}. Recent studies also explore self-supervised or large-scale pre-training tailored for medical images, showing promising results in reducing domain gaps and improving downstream tasks \cite{40Azizi2021}.
	
	\subsection{Pre-training in Medical Imaging}
	Transfer learning from ImageNet \cite{34Deng2009,28He2019} remains a popular strategy to overcome data scarcity, but the degree of improvement often depends on how closely medical images align with natural-image distributions \cite{27Raghu2019}. Indeed, certain studies have reported that, given a sufficiently large in-domain dataset, training from scratch can match or outperform ImageNet-pretrained models \cite{28He2019}. In the context of OCT imaging, the domain gap—arising from differences in acquisition physics, layer structures, and brightness distributions—can be substantial \cite{39Takahashi2024}. As summarized in Table~\ref{tab:relatedsummary}, such a gap may necessitate specialized model architectures or alternative pre-training strategies.
	
	Hence, whether ImageNet-pretrained Vision Transformers consistently yield better results on OCT classification tasks remains an open question. Although preliminary work suggests the benefits of domain-specific pre-training or self-supervised approaches \cite{41Ran2021}, no large-scale, publicly available ViT models pre-trained specifically on OCT data currently exist. This gap motivates our systematic comparison of \textbf{ViT (Pre-trained)} and \textbf{ViT (Scratch)} across different OCT dataset sizes and categories to clarify the role of pre-training in this specialized medical domain.
	\section{Materials and Methods}
	\label{sec:methods}
	
	\subsection{Dataset and Preprocessing}
	This study uses a publicly available OCT dataset introduced by Kermany et al. \cite{16Kermany2018}, which includes four categories of retinal pathologies: Choroidal Neovascularization (CNV), Diabetic Macular Edema (DME), Drusen, and Normal.
	
	\begin{figure}[H]
		\centering
		\includegraphics[width=0.65\textwidth]{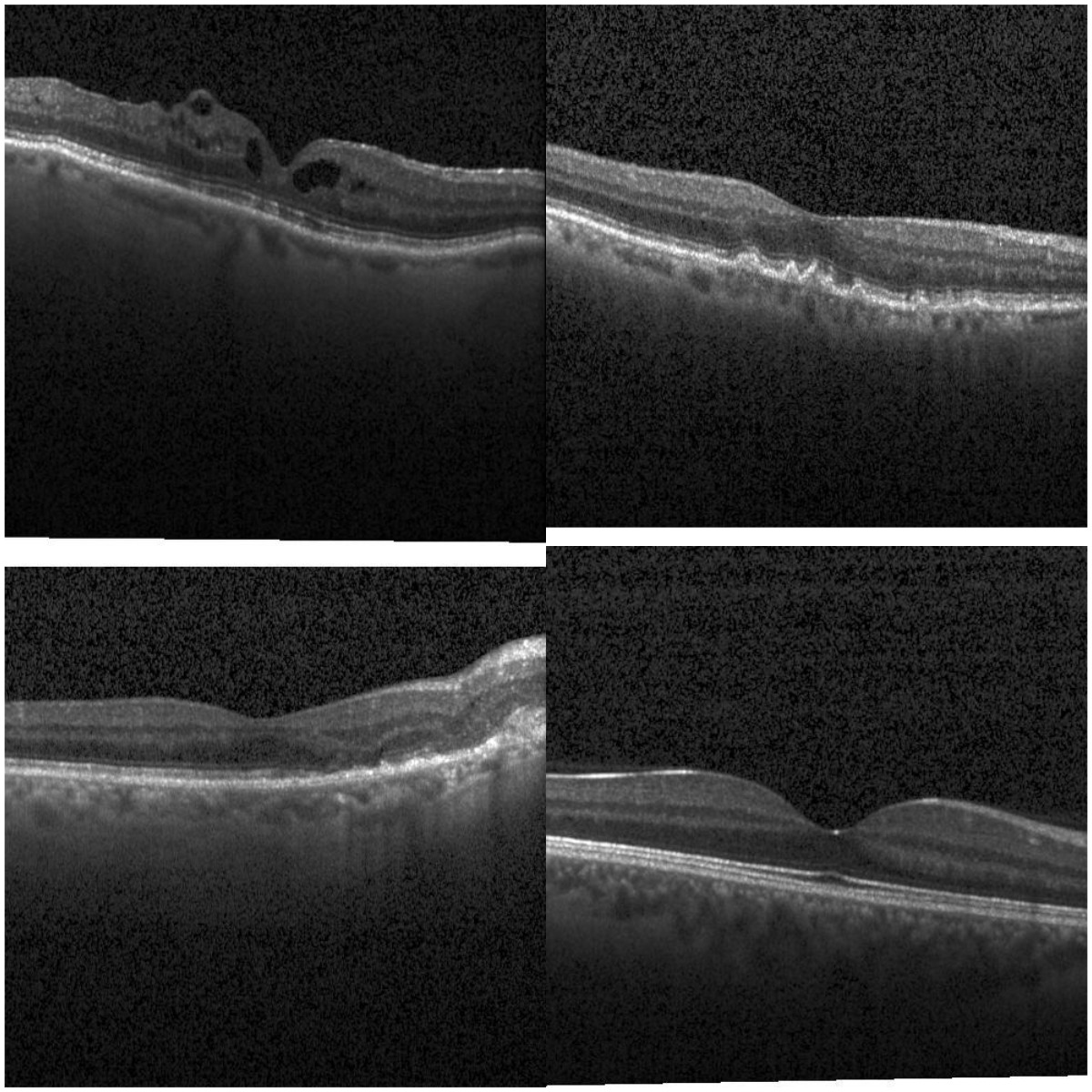}
		\caption{Examples of the four categories: clockwise from top-left—DME, Drusen, CNV, and Normal.}
		\label{fig:large_training}
	\end{figure}
	
	To evaluate model performance, we define two experimental settings:
	\begin{itemize}
		\item \textbf{Large-scale subset:} The training set contains over 2000 images, trained for 400 epochs, with a test set of approximately 400 images.
		\item \textbf{Small-scale subset:} The training set contains approximately 400 images, trained for 200 epochs, with a test set of about 100 images.
	\end{itemize}
	
	All images are resized or cropped to a standardized resolution (e.g., $224 \times 224$) while retaining the single-channel (grayscale) format characteristic of OCT images. Basic data augmentation, such as random flips and small rotations, is applied unless otherwise specified.
	
	\subsection{Implementation Details}
	
	\textbf{Vision Transformer Setup.} We adopt the ViT architecture described in \cite{10Dosovitskiy2020}, dividing images into $16\times16$ patches, each linearly projected into embeddings with position encodings. We compare two initialization strategies:
	\begin{itemize}
		\item \textbf{ViT (Scratch)}: All weights are randomly initialized.
		\item \textbf{ViT (Pre-trained)}: Initialized from an ImageNet21K-pretrained model, then fine-tuned on our OCT data.
	\end{itemize}
	
	\textbf{Training Protocol.} We use the Adam optimizer with a base learning rate of $1\times10^{-4}$, batch size 32, and train for up to 200--400 epochs depending on the dataset size. Early stopping or learning rate decay strategies are applied if validation accuracy saturates.
	
	\textbf{Evaluation Metrics.} We measure:
	\begin{itemize}
		\item \textbf{Accuracy (\%)}, \textbf{Loss} (cross-entropy)
		\item \textbf{Confusion matrices} for class-wise performance
		\item \textbf{ROC} (Receiver Operating Characteristic) curves and AUC for each class
	\end{itemize}
	
	\section{Experiments and Results}
	\label{sec:exp}
	
	We present the experimental findings under two data-scale settings. First, we show overall training/testing performance curves, followed by confusion matrices and ROC curves for in-depth class-wise analysis.
	
	\subsection{Training Curves}
	\subsubsection{Large-scale 4-class OCT Classification (400 epochs)}
	As summarized in Table~\ref{tab:oct4-large}, both \textbf{ViT (Pre-trained)} and \textbf{ViT (Scratch)} achieve high accuracy (around 90--91\%), with the scratch-trained model slightly edging out the pre-trained in the final stage.
	
	\begin{table}[H]
		\centering
		\caption{Performance on the 4-class OCT dataset (large-scale) after 400 epochs.}
		\label{tab:oct4-large}
		\begin{tabular}{lcc}
			\toprule
			\textbf{Method}     & \textbf{Test Accuracy (\%)} & \textbf{Final Test Loss}\\
			\midrule
			ViT (Pre-trained)   & $\approx 90.07\sim 90.86$   & 0.34--0.35 \\
			ViT (Scratch)       & $\approx 90.86\sim 91.00$   & 0.30--0.31 \\
			\bottomrule
		\end{tabular}
	\end{table}
	
	\begin{figure}[H]
		\centering
		\includegraphics[width=1\textwidth]{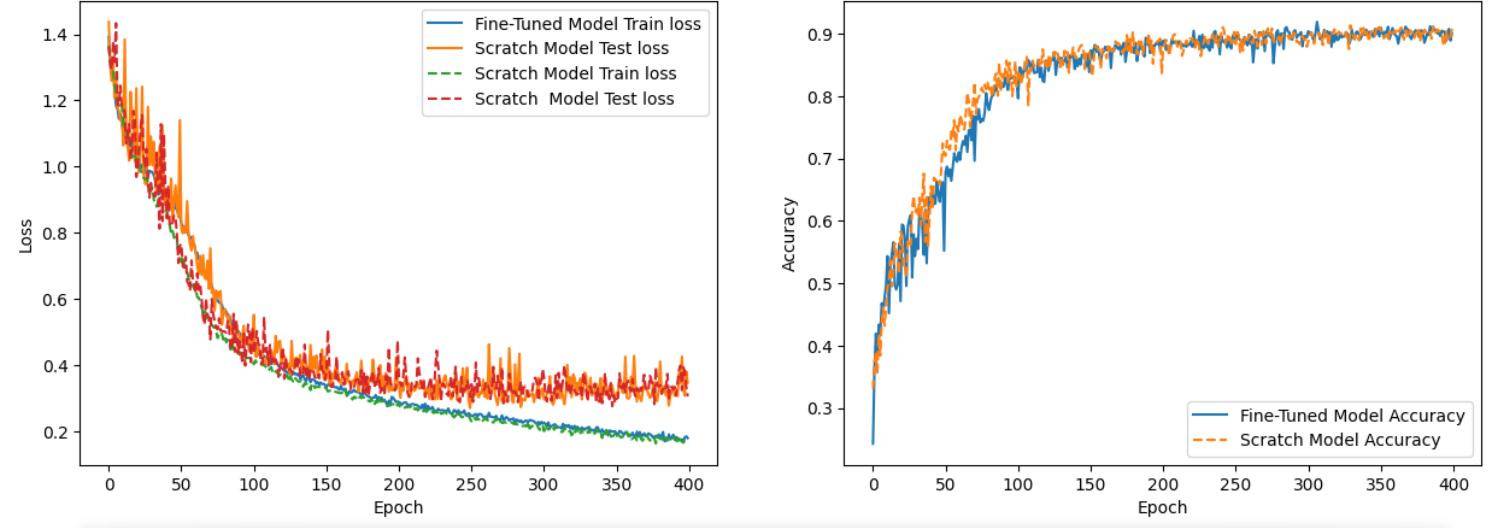}
		\caption{Training and test loss (left) and accuracy (right) for the large training set (2000+ images) over 400 epochs, comparing Fine-Tuned (Pre-trained) vs. Scratch ViT.}
		\label{fig:large_training2}
	\end{figure}
	
	\subsubsection{Small-scale 4-class OCT Classification (200 epochs)}
	In the reduced dataset (around 400 images), final accuracies are lower overall. Table~\ref{tab:oct4-small} shows that the scratch-trained model slightly outperforms the pre-trained one in final accuracy, although the latter converges faster in early epochs.
	
	\begin{table}[H]
		\centering
		\caption{Performance on the 4-class OCT dataset (small-scale) after 200 epochs.}
		\label{tab:oct4-small}
		\begin{tabular}{lcc}
			\toprule
			\textbf{Method}     & \textbf{Test Accuracy (\%)} & \textbf{Notes}\\
			\midrule
			ViT (Scratch)       & $\approx 60.35$             & -- \\
			ViT (Pre-trained)   & $\approx 57.36$             & Faster early convergence \\
			\bottomrule
		\end{tabular}
	\end{table}
	
	\begin{figure}[H]
		\centering
		\includegraphics[width=1\textwidth]{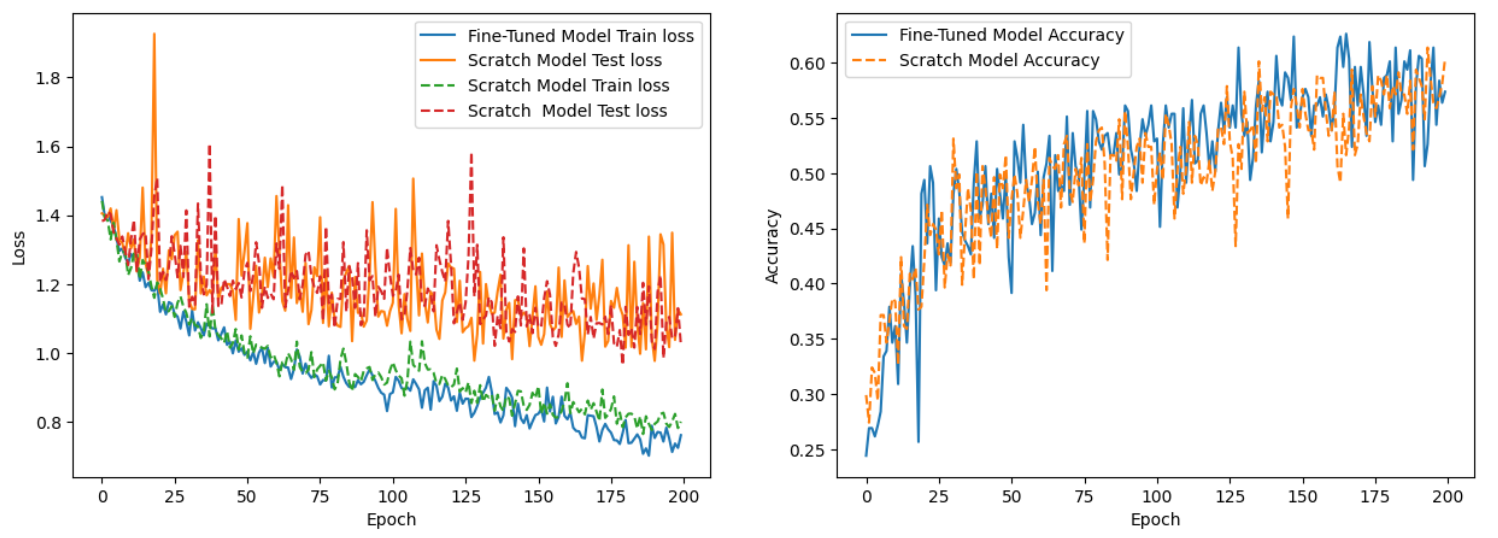}
		\caption{Training and test loss (left) and accuracy (right) for the smaller training set (about 400 images) over 200 epochs, comparing Fine-Tuned (Pre-trained) vs. Scratch ViT.}
		\label{fig:small_training}
	\end{figure}

	\subsection{Confusion Matrices}
	To further analyze class-wise predictions, we compare confusion matrices for \textbf{Pre-trained} vs. \textbf{Scratch} models under both large-scale (400 epochs) and small-scale (200 epochs) training conditions. 
    Each matrix entry represents the number (or percentage) of images from a true class (rows) classified as one of the predicted classes (columns).
	
	\subsubsection{Large-scale (400 epochs)}
	\begin{figure}[H]
		\centering
		\includegraphics[width=\textwidth, height=10cm, keepaspectratio]{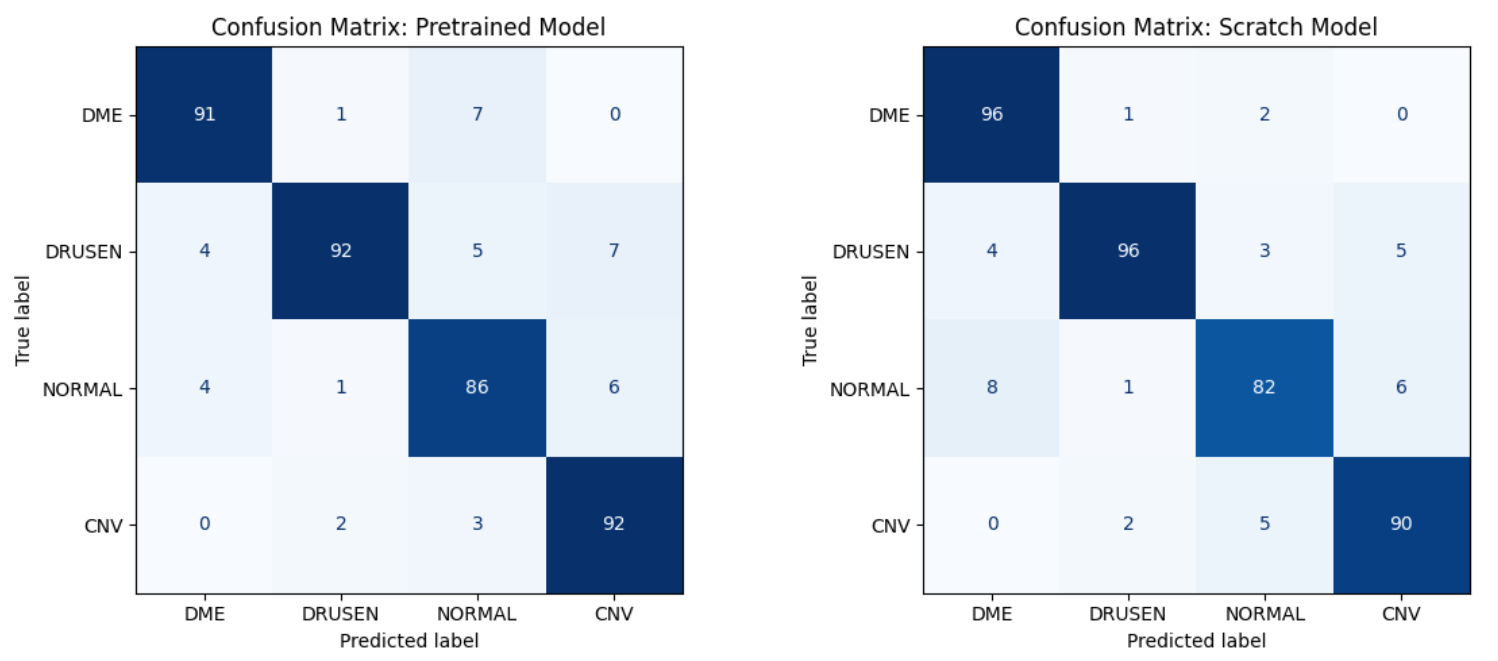} %
		\caption{Confusion Matrix for the large-scale dataset (400 epochs). 
			\textbf{Left:} Pre-trained model. \textbf{Right:} Scratch model.}
		\label{fig:confusion_large}
	\end{figure}
	
	\noindent
	From Figure~\ref{fig:confusion_large}, we observe that both models exhibit strong classification performance across all four classes (DME, DRUSEN, NORMAL, CNV), though minor differences can be seen in certain off-diagonal entries (e.g., Drusen vs. Normal misclassifications).
	
	\subsubsection{Small-scale (200 epochs)}
	\begin{figure}[H]
		\centering
		\includegraphics[width=\textwidth, height=10cm, keepaspectratio]{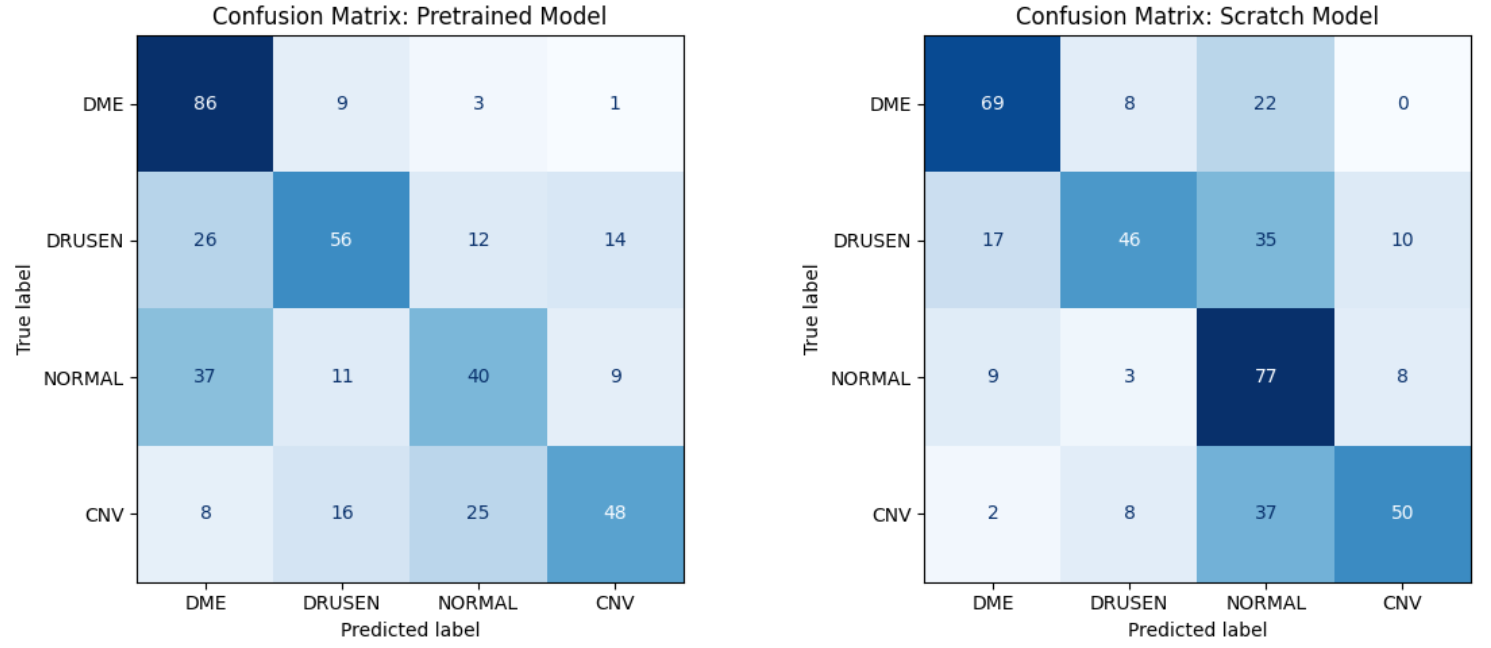}
		\caption{Confusion Matrix for the small-scale dataset (200 epochs). 
			\textbf{Left:} Pre-trained model. \textbf{Right:} Scratch model.}
		\label{fig:confusion_small}
	\end{figure}
	
	\noindent
	In the small-scale scenario (Figure~\ref{fig:confusion_small}), the confusion matrices reveal higher misclassification rates overall, reflecting the difficulty of learning robust features from limited data. Still, certain classes (e.g., DME) remain relatively well-predicted, while others (e.g., Normal vs. Drusen) show more confusion.
	
	\subsection{ROC Curves and AUC}
	Finally, we plot the ROC curves for each class (DME, DRUSEN, NORMAL, CNV) under both large-scale and small-scale settings to illustrate the true positive rate (TPR) vs. false positive rate (FPR) performance, along with the area under the curve (AUC).
	
	\subsubsection{Large-scale (400 epochs)}
	\begin{figure}[H]
		\centering

		\includegraphics[width=\textwidth, height=10cm, keepaspectratio]{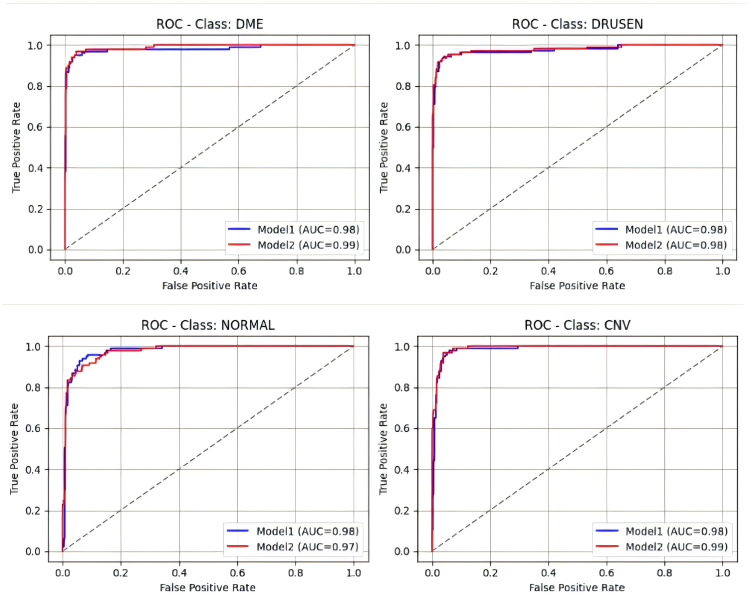}
		\caption{ROC Curves (by class) for the large-scale dataset (400 epochs), comparing Pre-trained (Model1) and Scratch (Model2).}
		\label{fig:roc_large}
	\end{figure}
	
	\noindent
	Figure~\ref{fig:roc_large} shows that both models achieve high AUC values (generally above 0.9 for most classes). This finding aligns with the confusion matrices in Figure~\ref{fig:confusion_large} indicating strong discriminative performance on the large dataset.
	
	\subsubsection{Small-scale (200 epochs)}
	\begin{figure}[H]
		\centering
		\includegraphics[width=\textwidth, height=10cm, keepaspectratio]{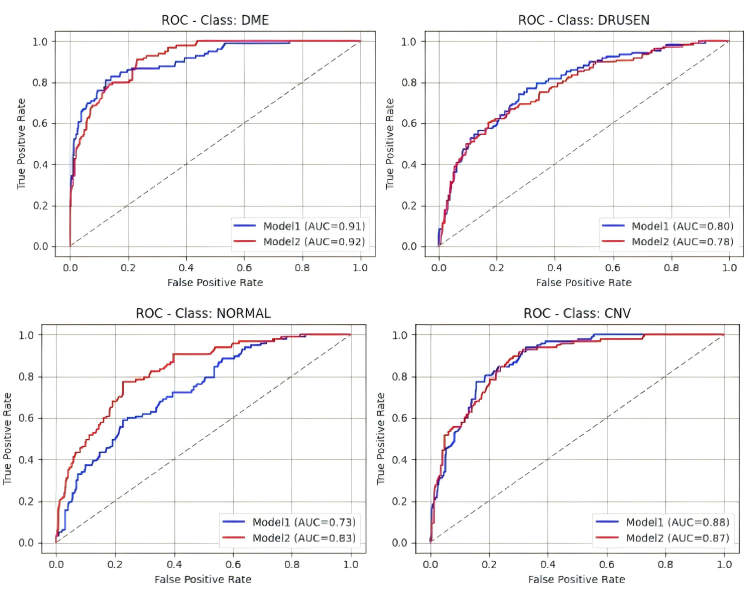}
		\caption{ROC Curves (by class) for the small-scale dataset (200 epochs), comparing Pre-trained (Model1) and Scratch (Model2).}
		\label{fig:roc_small}
	\end{figure}
	
	\noindent
	As illustrated in Figure~\ref{fig:roc_small}, the AUC values are generally lower than in the large-scale case, reflecting a more challenging classification scenario with limited data. Notably, the difference between Pre-trained vs. Scratch can be subtle in certain classes, suggesting that domain gap may limit the potential advantage of pre-trained weights.
	
	\section{Discussion}
	\label{sec:discussion}
	
	\subsection{Pre-trained vs. Scratch: Advantages and Limitations}
	From both the confusion matrices and ROC analyses, we see that:
	\begin{itemize}
		\item \textbf{Large-scale data (400 epochs)}: Pre-trained and Scratch models both excel, yielding high AUC and accurate confusion matrices (Fig.~\ref{fig:confusion_large}, \ref{fig:roc_large}). Scratch can match or slightly surpass Pre-trained in final accuracy.
		\item \textbf{Small-scale data (200 epochs)}: While Pre-trained converges faster, it does not ultimately outperform Scratch in terms of final metrics (Fig.~\ref{fig:confusion_small}, \ref{fig:roc_small}). 
	\end{itemize}
	These findings underscore that ImageNet-based pre-training, though beneficial for early-stage convergence, does not guarantee superior class-wise results if the dataset is sufficiently large or if the domain gap is substantial.
	
	\subsection{Impact of Data Scale and Domain Gap}
	A higher data volume reduces domain mismatch issues, allowing the model to learn discriminative OCT-specific features from scratch. Conversely, in a very limited setting, even the best pre-training might not fully adapt to the OCT domain. Potential improvements include using domain-specific pre-trained weights (if available) or leveraging advanced data augmentation to simulate greater variety.
	
	\section{Conclusion and Future Work}
	\label{sec:conclusion}
	
	We set out to examine whether \emph{a Pre-trained Vision Transformer is sufficient for OCT image classification}. Our results, based on training/test curves, confusion matrices, and ROC analyses, suggest:
	
	\begin{itemize}
		\item \textbf{Large-scale data:} Scratch vs. Pre-trained yield similarly high performance (AUC $>0.9$), with Scratch sometimes exceeding Pre-trained in final accuracy.
		\item \textbf{Small-scale data:} Pre-trained converges faster but does not consistently outperform Scratch by the end, partly due to domain gap and limited data.
		\item \textbf{Class-wise analysis:} Both approaches handle certain categories (e.g., DME) better, while others (e.g., Normal vs. Drusen) exhibit more confusion, especially in small-scale conditions.
	\end{itemize}
	
	In conclusion, a pre-trained ViT is not strictly necessary for strong classification performance in OCT tasks—particularly under a sufficiently large dataset. Nonetheless, pre-training may expedite early-stage convergence in data-scarce scenarios. Future work may explore creating large-scale OCT pre-training corpora or leveraging self-supervised paradigms to reduce reliance on ImageNet-based weights.
	
	\section*{Ethical Statement}
	\addcontentsline{toc}{section}{Ethical Statement}
	
	 Retina Images for DME and Drusen: These images are publicly available and were sourced from Kaggle, based on \cite{16Kermany2018}. All necessary ethical standards and patient consent were secured and adhered to.
	
	\clearpage
	
	\bibliographystyle{plain}
	\bibliography{Refer}
	
\end{document}